\PassOptionsToPackage{table,dvipsnames}{xcolor}
\documentclass[letterpaper]{article} 
\usepackage{aaai2026}  
\usepackage{times}  
\usepackage{helvet}  
\usepackage{courier}  
\usepackage[hyphens]{url}  
\usepackage{graphicx} 
\usepackage{natbib}  
\usepackage{caption} 
\usepackage{algorithm}
\usepackage{algorithmic}
\usepackage{siunitx}
\usepackage{newfloat}
\usepackage{listings}
\usepackage{multirow}

\DeclareCaptionStyle{ruled}{labelfont=normalfont,labelsep=colon,strut=off} 
\floatstyle{ruled}
\newfloat{listing}{tb}{lst}{}
\floatname{listing}{Listing}
\usepackage{multirow}
\usepackage{array}
\usepackage{nicematrix}
\usepackage{booktabs}

\usepackage{enumitem}
\usepackage{tabularx}
\usepackage{xspace}
\usepackage{amsmath, amssymb}
\usepackage{arydshln}
\usepackage{inconsolata}
\usepackage{soul}
\usepackage{makecell}
\usepackage{listings}
\usepackage{caption,subcaption}
\usepackage{soul}
\usepackage{anyfontsize}
\usepackage{tcolorbox}
\usepackage{changepage} 
\usepackage{csquotes}   

\newcommand{\system}[1]{\textsc{#1}\xspace}  
\newcommand{\EQSUMM}{\system{ArQuSumm}}

\title{\EQSUMM: Argument-aware Quantitative Summarization \\ of Online Conversations}

\author{
    An Quang Tang,
    Xiuzhen Zhang~\thanks{Corresponding author.},
    Minh Ngoc Dinh,
    Zhuang Li
}

\affiliations{
    RMIT University, Australia\\
    s3695273@rmit.edu.vn, xiuzhen.zhang@rmit.edu.au, \\ minh.dinh4@rmit.edu.vn, zhuang.li@rmit.edu.au
}

\graphicspath{ {./images/} }

\begin{document}

\maketitle

\begin{abstract}
Online conversations have become more prevalent on public discussion platforms (e.g. Reddit).
With growing controversial topics, it is desirable to summarize not only diverse arguments, but also their rationale and justification.  
Early studies on text summarization focus on capturing general salient information in source documents, overlooking the argumentative nature of online conversations.
Recent research on conversation summarization although considers the argumentative relationship among sentences, fail to explicate deeper argument structure within sentences for summarization.
In this paper, we propose a novel task of argument-aware quantitative summarization to reveal the claim-reason structure of arguments in conversations, with quantities measuring argument strength.
We further propose \EQSUMM, a novel framework to address the task.  
To reveal the underlying argument structure within sentences, \EQSUMM leverages LLM few-shot learning grounded in the argumentation theory to identify propositions within sentences and their claim-reason relationships.  
For quantitative summarization, \EQSUMM employs argument structure-aware clustering algorithms to aggregate arguments and quantify their support. 
Experiments show that \EQSUMM outperforms existing conversation and quantitative summarization models 
and generate summaries representing argument structures that are more helpful to users, of high textual quality and quantification accuracy.

\end{abstract}

\begin{links}
    \link{Code}{https://github.com/antangrocket1312/ArQuSumm}
\end{links}

\section{Introduction}
The proliferation of user participation of online conversations platforms such as online discussion forums
(e.g.,Reddit~\footnote{\url{https://www.reddit.com/}})~\citep{volske-etal-2017-tl,zhang2019reading}, 
community question answering sites and online news discussion
forums has resulted in large volumes of online conversation data.
Summarization of online converstations has become an important text summarization task~\citep{fabbri-etal-2021-convosumm}.
Different from well structured documents such as news articles or scientific papers, online conversations,  
with many users frequently include hundreds of arguments, can spread across the various threads of online conversations~\citep{syed-etal-2023-frame}.
In fact, users not only engage in discussions to express their viewpoints, but also debate, justify, and challenge 
others' viewpoints. 
User comments are therefore often argumentative, with inferential relations among propositions expressing not only what people believe, but also why.

\begin{figure}[tbh]
    \centering
    \includegraphics[width=0.48\textwidth]{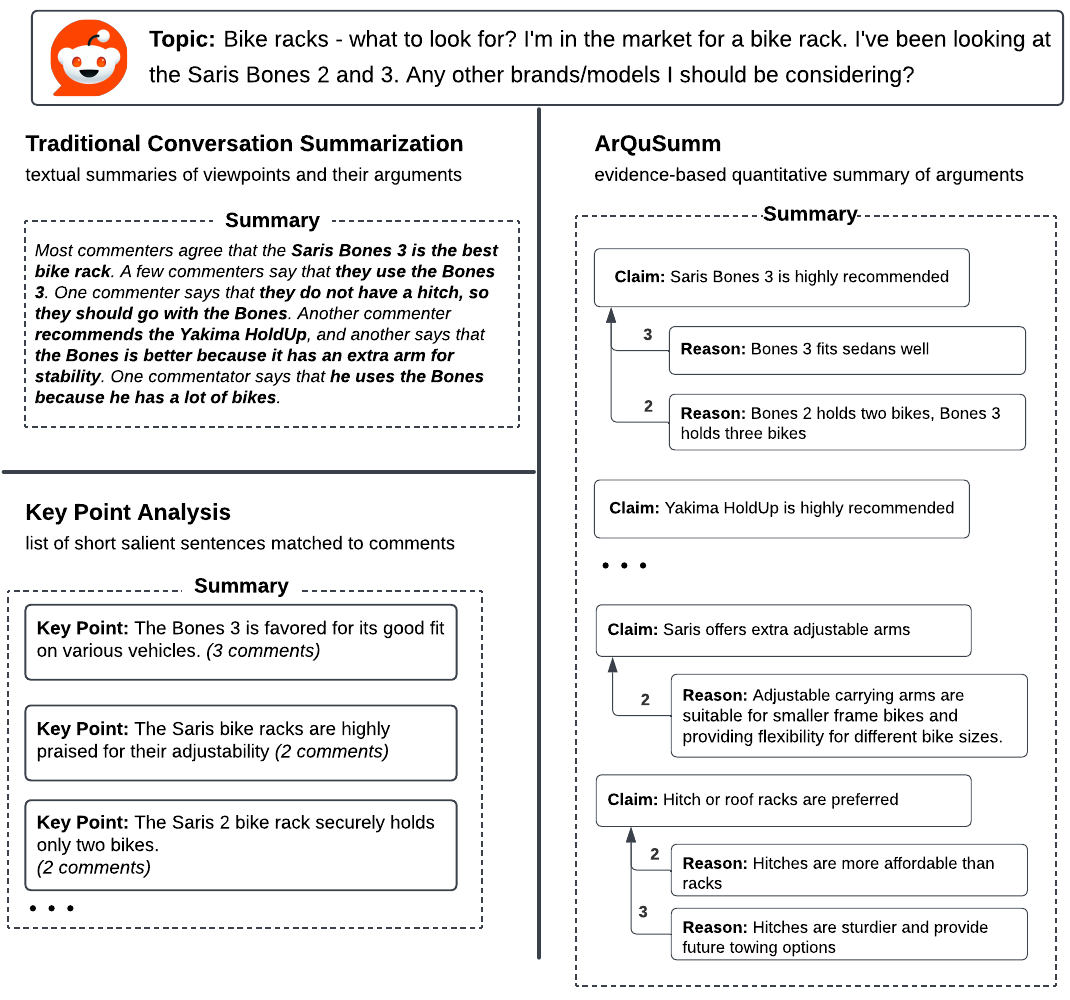}
    \caption{Comparison of \EQSUMM. and existing conversation and quantitative summarization methods}
    \label{fig:QFQS_Overview}
\end{figure}

Conventional multi-document textual summarization aims to output the most salient parts of multiple related documents in a concise and readable form.
Early research on conversation summarization utilized document summarization models to summarize conversations~\citep{liu-lapata-2019-text,lewis-etal-2020-bart,zhang2020pegasus}, 
but success is limited 
because conversational text contains main points scattering across multiple utterances and between numerous writers~\citep{gliwa-etal-2019-samsum}.
Recent studies then adopted argumentation theories~\citep{barker-gaizauskas-2016-summarizing} and frameworks~\citep{barker-etal-2016-sensei}
to model arguments and viewpoints presented in the conversation for summarization~\citep{ lenz2020towards,chen-yang-2021-structure,fabbri-etal-2021-convosumm}.
Still, existing studies fail to capture and comprehensively present the rationale and justifcation for arguments in their plain textual summary.
Further, arguments are modelled at the sentence level where sentences are the argument elements, often resulting in redundancy and incoherence in argumentative logic across sentences.
Importantly, the plain summary text lacks the ability to explicitly represent the argument structure and explain the reasons for claims.  

Recently the quantitative view was introduced into textual summaries to capture and numerically quantify diverse viewpoints in 
reviews and debates, 
in a task known as Key Point Analysis (KPA)~\citep{bar-haim-etal-2020-arguments,bar-haim-etal-2020-quantitative,bar-haim-etal-2021-every,2024aspectbased,tang-etal-2024-prompted}.
KPA summarizes user comments (sentences) into concise sentences called key points (KPs), which are claims expressing user viewpoints, and quantify their prevalence.
Nevertheless, KPA only generate a summary as a list of claims without capturing the argumentive logic between them, leaving to users to assess the reliability of these KPs on their own.
Moreover, the quantification of prevalence for KPs is mostly via matching KPs with users comments based on only semantic similarity in texts, which can be inaccurate. 

In this paper, we propose a novel framework \EQSUMM for argument-aware quantitative summarization of conversations. 
Different from prior studies, the summary includes \emph{claim-reason argument structures} explaining arguments in the summary.
Figure~\ref{fig:QFQS_Overview} shows an example. 
When discussing 
``which bike rack models to look for",
\EQSUMM presents every argument directly as a tree structure; the root represents the claim from user comments (e.g. \emph{should use Saris Bone 3}), and each leaf represents a reason for the claim (e.g., \emph{Bone 3 fits sedan well}), along with its support quantity (number of comments).

Specifically, to explicate the argument structure in user comments, the \EQSUMM framework first leverages LLM in-context learning grounded in argumentation theory to identify propositions as argument elements within sentences.
It then employs LLM in-context learning to predict the entailment relationship and identify claim and reason for the structure of arguments. 
Importantly, we propose an  argument structure-aware clustering algorithms to aggregate fine-grained propositions into distinct high-level claims and quantify their supporting reasons effectively.

Our main contributions are:
\begin{itemize}
    \item We introduce the novel task of argument-aware quantitative summarization of online conversations, where, unlike conventional plain textual summaries, argument structures are explicitly represented. 
    Experiments show that our new form of summary offers 7 times more comprehensive and useful presentation, and our proposed framework 
    outperforms baselines with up to 
    3.71 times improvement in textual similarity with ground-truth arguments and up to 0.421 improvement in F1 matching over 
    current quantitative summarization 
    framework~\citep{tang-etal-2024-prompted}.
    \item Different from existing studies we leverage LLM in-context learning grounded in argumentation theory to identify text spans of argument elements in sentences. 
    \item Leveraging LLM in-context learning to identify claim-reason relations for arguments, we design novel argument structure-aware clustering algorithms to form and quantify arguments and generate the structured summary. 
\end{itemize}

\section{Related Work}

\paragraph{Textual Summarization of Conversations}
Existing studies focus on textual summarization of conversations, utilizing the conversational structure.
\citet{barker-etal-2016-sensei} proposed conversation overview summary to capture the key contents of reader comment conversations for news articles. 
\citet{misra2017using} use summarization to discover central propositions in online debates.
\citet{barker-gaizauskas-2016-summarizing} identify three key components of conversational dialogues to manually construct an argument graph for the whole conversation.
Building on this theoretical  framework for argumentation, 
\citet{fabbri-etal-2021-convosumm} applied entailment relations to automatically construct argument graphs for conversations and generate textual summaries. 
Note that these existing studies model argument relations at the sentence level, which is a loose assumption according to the argumentation theory~\cite{toulmin1958uses}.

\paragraph{Quantitative Summarization of Key Points}
Key Point Analysis (KPA) was recently proposed to summarize key points (KPs) quantify them for reviews and debates~\citep{bar-haim-etal-2020-arguments,bar-haim-etal-2020-quantitative}.
To address that a flat list of KPs often expresses related ideas at varying levels of granularity, \citet{cattan-etal-2023-key} proposes to summarie key points into a hierarchy.
But their pipeline approach of first generating and then summarizing KPs can give rise to cascading of errors, leading to poor KP hierarchies.

\paragraph{Argument Mining}
Recent argument mining studies focus on identifying argumentative units and structures based on argumentation theories~\citep{stab-gurevych-2014-identifying}.
\citet{gupta-etal-2024-harnessing} proposes a quantitative argumentation framework (QAF) that harnesses LLMs and Toulmin theory~\citep{toulmin1958uses} to explicate and cluster arguments from comments into a hypergraph.
However, these graphs lack conciseness and is cluttered with multiple layers of arguments,
with some being possibly overlapping due to surface-level argument clustering approach.
Still, existing studies can only produce an argument graph to visualize the argument flow rather than a concise and readable summary.

\section{Task Formulation}
\label{sec:task}
Let \mbox{$\mathcal{D} = \{d_i\}_{i=1}^{|\mathcal{D}|}$} denote a collection of input comments in an online discussion thread.
From each comment $d_i$, we extract a set of argument propositions, namely a claim $c$ and its associated reasons $\{r_j\}$. This results in a set of extracted propositions for $\mathcal{D}$, denoted as $\mathcal{P} = \{(c_i, \mathcal{R}_i)\}_{i=1}^{|\mathcal{P}|}$, where each $c_i$ is a claim and $\mathcal{R}_i$ is a set of reasons justifying it.
In the the Argument-aware Quantitative Summarization (\EQSUMM) task, we aim generate a structured summary $\mathcal{S}$, which consists of a set of \textit{claims} $\hat{C} = \{\hat{c}_k\}_{k=1}^K$, each supported by a set of aggregated \textit{reasons} $\hat{\mathcal{R}}_k = \{\hat{r}_{k1}, \hat{r}_{k2}, \ldots\}$.
Formally, we define our summary as:
\[
\mathcal{S} = \left\{\left(\hat{c}_1, \hat{\mathcal{R}}_1\right), \left(\hat{c}_2, \hat{\mathcal{R}}_2\right), \ldots, \left(\hat{c}_k, \hat{\mathcal{R}}_k\right) \right\}_{k=1}^{K}
\]
where each $\hat{c}_k$ represents a semantically unified cluster of input claims from $\mathcal{P}$, and each $\hat{r}_{kj}$ corresponds to a distinct subgroup of reasons 
drawn from the reason sets $\mathcal{R}_i$ 
originally linked to the clustered claims.
For simplicity, hereafter we refer to $\hat{c}_k$ and $\hat{\mathcal{R}}_k$ produced in the final structured summary as \emph{claim} and \emph{reason}.
For example, a final summary entry may take the form: \\
\textbf{Claim:} \textit{Remote work improves productivity} \\
\hspace*{0.1em} $\rightarrow$ \textbf{Reason:} \textit{Less time is wasted on commuting} (22 instances) \\
\hspace*{0.1em} $\rightarrow$ \textbf{Reason:} \textit{Home environment allows for fewer distractions} (14 instances)

\section{The \EQSUMM framework}
\begin{figure*}[tbh]
    \centering
    \includegraphics[width=1\textwidth]
    {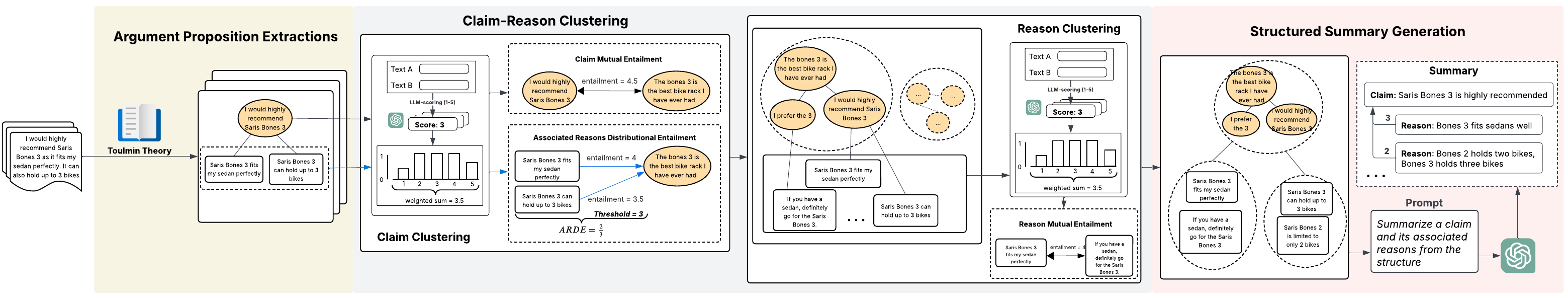}
    \caption{The \EQSUMM framework.
    }
    \label{fig:ArQuSumm_Overview}
\end{figure*}

Figure~\ref{fig:ArQuSumm_Overview} illustrates the overall pipeline of our proposed \EQSUMM framework. Given a set of comments from an online discussion.
\EQSUMM performs argument-aware quantitative summarization by generating a structured summary of claims and their supporting reasons, each annotated with prevalence. The framework consists of three stages: 
\emph{i) Argument Proposition Extraction}, 
\emph{ii) Claim-Reason Clustering}, and 
\emph{iii) Structured Summary Generation}.

\subsection{Argument Proposition Extraction}
Unlike previous summarization studies, we ground our summarization process on claims and reasons, i.e., text phrases, implied by the comments.
The process is more effective and faithful than summarizing and quantifying on comment sentences because it can (1) bypass noise in the original sentence, (2) recover the original entity name in case of cross-sentence references (e.g., John instead of he).
We specifically harnesses the LLM's extensive knowledge on argumentation theory, by zero-shot prompting it to extract possible argument components inside a comment. 
Based on the experiment of various theories~\citep{toulmin1958uses, walton1996argumentation, freeman1991dialectics} for argument explication, we decided to make use of Toulmin~\citep{toulmin1958uses} due to LLM's exceptional interpretation and capability to extract arguments following this theory~\citep{gupta-etal-2024-harnessing}.
In particular, we prompt an LLM with references to Toulmin’s theory (e.g.,
‘According to Toulmin model,’)~\citep{gupta-etal-2024-harnessing}, 
which elicits a response that correctly bases on Toulmin’s theory to extract values from the input comment corresponding to the 'claim' and 'reason' from the theory.
The Toulmin theory decomposes the arguments within a comment into three components, namely:
\textbf{\textit{claim}} (assertion or viewpoint made by the author for general acceptance), 
\textbf{\textit{reason}} (proposition provided by the author to convince the audience to accept the claim), and 
\textbf{\textit{warrant}} (the author’s world knowledge explain why the claim follow from the provided reason).
Note that we omit the warrant layer, generated by LLM’s parametric knowledge, to preserve original user reasoning.
Note also that arguments in a comment (with multiple sentences) can carry multiple claims, where each claim could be supported by multiple reasons.

\subsection{Claim-Reason Clustering}
Previous approaches~\citep{gupta-etal-2024-harnessing} clustered argument propositions based on their embeddings irrespective of their role in the argument structure, which impairs the alignment between claims and their associated reasons and fails to explicitly preserve the inferential relationship between them.
A key innovation of \EQSUMM lies in disentangling and separately clustering claims and reasons to better reflect the underlying argument structure.
We therefore propose a two-stage hierarchical clustering process grounded in argumentation theory and entailment-based reasoning.

\subsubsection{Claim Clustering}
At the highest level of social discussion, identifying distinct claims to represent diverse viewpoints is central to argument summarization.
Prior approaches often rely on semantic similarity of claims and reasons--typically using cosine distance in embedding space--to induce viewpoint clusters~\citep{gupta-etal-2024-harnessing}. However, such surface-level similarity overlooks critical differences in attitude, stance, and implied judgment that distinguish opposing viewpoints.
In this work, we argue that clustering based solely on claim embeddings is insufficient for capturing meaningful argumentative distinctions. Instead, we propose leveraging entailment relationships between argument propositions as a proxy for viewpoint alignment.

\paragraph*{Associated Reasons Distributional Entailment}
Clustering claims solely based on claim's mutual entailment score might be ineffective and unreliable, 
while highly relevant claims might still have associated reasons unsupportive to each other.
To mitigate this risk, we propose incorporating information from each claim's associated reasons as a regularizing signal in the clustering process.
Our method draws inspiration from the distributional inclusion hypothesis~\citep{geffet-dagan-2005-distributional},
which suggests that the context surrounding an entailing word $w_1$ is naturally expected to occur also with the entailed word $w_2$~\citep{geffet-dagan-2004-feature}.
We adapt this hypothesis to the argumentative domain with the intuition that
if claim $c_{m}$ supports claim $c_{n}$, it is likely that a reason $r_j \in \mathcal{R}_{m}$ that supports $c_{m}$ will also support $c_{n}$.  
We formalize this as the \textbf{Associated Reasons Distributional Entailment (ARDE)} score.
Given a claim \( c_m \) and its associated reasons \( \mathcal{R}_m \), we compute the proportion of reasons that also support a neighbouring claim \( c_n \) as:
\[
\text{ARDE}(c_m \Rightarrow c_n) = \frac{|\{ r_j \in \mathcal{R}_m : r_j \text{ supports } c_n \}|}{|\mathcal{R}_m|}
\]
where a reason $r_j$ of claim $c_{m}$ is determined to support adjacent claim $c_{n}$ if it exceeds a threshold $t$.

To cluster claims into distinctive groups, we construct a \textit{fully connected graph} by computing two types of pairwise scores between all claims \( c_i \in \mathcal{P} \): (1) the \textit{entailment score}, and (2) the ARDE score between claim pairs in both directions.
These scores are then combined to build a undirected graph \( G_{claim}\), where nodes represent claims and edges capture mutual support relationships. Specifically, for each candidate pair \( (c_m, c_n) \), we compute a final alignment score \( s(m, n) \) as the average of the \textit{bidirectional entailment scores} and \textit{bidirectional ARDE scores}~\footnote{we average the pairwise score from each direction to obtain the bidirectional score}.
\[
s(m, n) = \frac{1}{2} \left( \text{Entail}_{\text{bi}}(c_m, c_n) + \text{ARDE}_{\text{bi}}(c_m, c_n) \right)
\]
An edge \( e(m, n) \) is added to \( G_{claim} \) if \( s(m, n) > \tau \), where \( \tau \) is a threshold controlling clustering granularity. 
Claims within the same \textit{strongly connected component} of \( G \) are contracted into the same cluster, as they are mutually supportive in both semantic content and reasoning structure.

\subsubsection{Aggregation and Clustering of Reasons}
After forming clusters of semantically equivalent claims, we aggregate their associated reasons to form a set of supporting evidence for each claim group. Specifically, for a claim cluster $\hat{c}_k$, we define its aggregated reason set as:
\(
\hat{\mathcal{R}}_k = \bigcup_{c_i \in \text{cluster}_k} \mathcal{R}_i
\)
Next, we perform \textit{reason clustering} within each $\hat{\mathcal{R}}_k$ to identify distinct subgroups of justifications. 
Each resulting subgroup represents a coherent reasoning pattern frequently cited in support of the aggregated claim. 
To determine whether two reasons $r_u, r_v \in \hat{\mathcal{R}}_k$ belong to the same cluster, we build a bipartite entailment graph to compute their \textit{pairwise entailment scores} in both directions. 
Based on these scores, 
we construct an undirected graph $G = (\hat{\mathcal{R}}_k, E)$, where each node corresponds to a reason within $\hat{\mathcal{R}}_k$, and an undirected edge $\{r_u, r_v\} \in E$ is added if and only if average entailment score from both direction exceed a threshold $\tau$.
We then contract each strongly connected component of $G$ to be a \textit{reason cluster}, which represents a distinct subgroup of justification commonly used to support claim $\hat{c}_k$ from online discussion.

\paragraph{LLM-based Entailment Scoring Function}
Recent studies suggest using LLMs as reference-free metrics for NLG evaluation.
Inpsired by the \texttt{G-Eval} LLM-based evaluator~\citep{liu-etal-2023-g} for NLG output,
we adapt this evaluator to the Natural Language Inference (NLI) task for scoring the entailment relationship  between argument propositions extracted from social comments.
Specifically, we utilize the probabilities of output tokens from LLMs to normalize the scores and take their weighted summation as the final results. Formally, given a set of scores (like from 1 to 5) predefined in the prompt 
$S=\{s_1, s_2, ..., s_n\}$, the probability of each score $p(s_i)$ is calculated by the LLM, and the final entailment score is:
\begin{equation}
    score = \sum_{i=1}^n p(s_i) \times s_i
\end{equation}
The intuition is prompting the LLM to score the two given texts multiple times and then take the average of all runs to achieve fine-grained, continuous scores that better reflect the entailment relationship between argument propositions.
Empirical validation shows that our LLM-based entailment scoring (using \texttt{GPT-4.1}) obtain stronger alignment with human judgement than scores produced by the state-of-the-art \texttt{RoBERTa-large} model~\footnote{\url{https://huggingface.co/roberta-large-mnli}} ($r$ = 0.556, Accuracy = 0.715)\footnote{Details in Supplementary Material}.

\subsection{Structured Summary Generation}
Given the final set of claim clusters $\{\hat{c}_k\}_{k=1}^K$ and their corresponding reason clusters $\{\hat{\mathcal{R}}_k\}_{k=1}^K$, we generate the bullet-like structured summary $\mathcal{S}$ 
where each bullet is a tree structure, with its stem node representing $\hat{c}_k$ and the connected leaf nodes representing $\hat{\mathcal{R}}_k$.
Note that due to structural complexity and to ensure highly-coherent content, we prompt an LLM to summarize each claim cluster and its corresponding reason clusters for each argument in the discussion per generation, before aggregating output across all arguments to achieve a final claim-reason structure summary.
Note also that to minimize ambiguity and hallucination, we explicitly structure the input of $\hat{c}_k$ and $\hat{\mathcal{R}}_k$ as a JSON object.
Importantly, we assign each claim and reason group with a unique identifier at input, and enforce the LM to include reference these IDs at output to ensure accurate alignment between summarized claims and reasons and their source clusters.

\paragraph{Prompt Engineering}
Following best practices from OpenAI's prompt engineering guidelines\footnote{\url{https://platform.openai.com/docs/guides/prompt-engineering}}, we design the prompt with three key components:
(1) a brief task description explaining the summarization goal and input format, (2) a clear instruction to generate a general claim and distinct reasons with a maximum length of 10 tokens, (3) step-by-step guidelines for how the LLM should transform the input 
(i.e., infer a concise claim from the claim list, then derive specific, supporting ground key points from each reason cluster).

\section{Experiments \& Results}
\subsection{Datasets and Experiment Setting}
We evaluate our framework on ConvoSumm~\citep{fabbri-etal-2021-convosumm}, an abstractive conversation summarization corpus with diverse conversation datasets (domains),
but specifically focus on three datasets:
\textbf{NYT} (New York Times news comments), \textbf{Reddit} (discussion forums), and \textbf{Stack} (StackExchange community question answering), since they are most relevant to the argument-rich discussions targeted by \EQSUMM.
Each instance includes user comments from multi-turn threads, and a crowd-sourced abstractive summary capturing diverse viewpoints.
We use only the test split for evaluation, which contains 250 examples per dataset.

We experimented \EQSUMM with \texttt{GPT-4.1}, and \texttt{Mistral-7B}\footnote{\url{https://huggingface.co/mistralai/Mistral-7B-Instruct-v0.3}} as backbone LLMs.
For our LLM-based entailment scoring, we
sample 5 times to estimate the weighted summation of entailment score of a given claim or reason pair.
Note that for runtime and cost feasibility, for all clustering stages, we batched and performed LLM-based Entailment scoring in an one-to-many manner 
(e.g., scoring 1 claim vs a list of other claims/reasons per prompt)
instead of pairwise.
We set the clustering threshold \(\tau = 3\) based on empirical inspection of the cluster quality.

\subsection{Baselines}
We benchmark \EQSUMM against a various baselines, covering existing conversation summarization and quantitative summarization (KPA) framework.
\paragraph{ConvoSumm} 
A conversation summarization framework
that integrates argument mining with abstractive summarization~\citep{fabbri-etal-2021-convosumm}.
Sentences from a comment were first classified as claims or premises, i.e., reasons, before being mapped for relations within and across comments using a RoBERTa~\citep{liu2019roberta} model finetuned on  MNLI entailment 
\footnote{\url{https://huggingface.co/FacebookAI/roberta-large-mnli}} to construct an argument graph.
Graph-based information are then linearized as input to a BART-large~\citep{lewis-etal-2020-bart} abstractive summarization model fine-tuned for graph-to-text generation.

\paragraph{\texttt{GPT-4.1}-ICL}
We few-shot prompt (with two in-context examples) a \texttt{GPT-4.1} model, as an end-to-end solution, to directly process a list of argument propositions and output a structured summary.
The prompt adopts the Chain-of-Thoughts strategy, which guides and elicits the LLM to generate output with four reasoning steps: (1) grouping equivalent claims, (2) aggregating associated reasons, (3) clustering similar justifications, and (4) summarizing each claim with its reason clusters.

\paragraph{QAF}
An adapted version of the quantitative argumentation framework (QAF) of \citet{gupta-etal-2024-harnessing} for the \EQSUMM task, 
which basically clusters argument propositions without distinguishing their roles (e.g., claim or reason). It follows four stages: (1) extract propositions from comments via LLM prompting, (2) embed and cluster them using Sentence-BERT and DP-means, (3) infer directed edges to recover reason-to-claim links for hypergraph construction, and (4) generate structured summaries from clusters in the first two hypergraph levels.

\paragraph{ConvPAKPA/ConvRKPA}
Adapted versions of the KPA review summarization systems PAKPA~\citep{tang-etal-2024-prompted} and RKPA~\citep{bar-haim-etal-2021-every} for conversation summarization, using extracted argument propositions instead of comment sentences. \textbf{ConvPAKPA} identifies aspects and associated sentiment, clusters by aspect–sentiment pairs, and prompts LLMs to generate aspect-specific key points. \textbf{ConvRKPA} uses a quality ranking model to select KP candidates, then matches propositions using a KP Matching model~\citep{bar-haim-etal-2020-quantitative}.

We conducted comprehensive evaluation of our \EQSUMM framework along the dimensions of argument quality, textual quality of arguments, as well as quantification accuracy. In addition to leveraging LLMs for automatic evaluation, we also employ human evaluation. We organised the results into sections based on research questions.

\begin{table}[htbp]
    \scalebox{0.6}
    {
    \begin{tabularx}{0.8\textwidth}{c:lccc:cccc:c}
        \toprule
        Summary & Baseline & CV & FF & RD & VL & SN & IN & SA & HF\\
        \midrule
        \multirow{3}{*}{\makecell{Structured}} & \EQSUMM & \textbf{26.51} & \textbf{25.34} & \textbf{21.39} & \textbf{31.85} & \textbf{37.41} & \textbf{31.95} & \textbf{23.19} & 54.46 \\
         & \texttt{GPT-4.1}-ICL & 22.17 & 24.54 & 20.18 & 30.53 & 24.43 & 24.05 & 21.24 \\
        & QAF & 13.60 & 16.74 & 18.53 & 13.23 & 15.23 & 20.36 & 18.94 \\
        \hdashline
        \multirow{1}{*}{Textual} & ConvoSumm & 13.61 & 14.49 & 16.56 & 12.83 & 07.73 & 11.20 & 13.10 & \multirow{1}{*}{36.72} \\
        \hdashline
        \multirow{2}{*}{KP} & ConvPAKPA & 14.83 & 11.46 & 14.36 & 06.98 & 10.43 & 09.02 & 13.47 & 08.83 \\
        & ConvRKPA & 09.28 & 07.44 & 08.98 & 04.58 & 04.76 & 03.43 & 10.06 \\
        \bottomrule
    \end{tabularx}
    }
    \caption{Human evaluation of summary's information quality.
    Reported are the Bradley Terry scores of 8 dimensions, from left to right, \textsc{Coverage}, \textsc{Faithfulness} and \textsc{Redundancy}, \textsc{Validity}, \textsc{Sentiment}, \textsc{Informativeness}, \textsc{SingleAspect}, \textsc{Helpfulness}.
\label{table:kp_quality_eval_results_T}}
\end{table}

\subsection{RQ1: Is the argument structure in the summaries more helpful to users?}
\paragraph{Settings}
We manually evaluate both \emph{the utility of the argument structure} in the summaries, and also the summary's \emph{information quality} of different baselines, using a set of 8 evalation dimension.
While information quality is adopted from the 7 different dimensions, 
(e.g., \textsc{Faithfulness}, \textsc{Coverage})
defined by previous KPA studies~\citep{kapadnis-etal-2021-team} (see Appendix), 
we additionally define \textsc{Helpfulness} to evaluate `` how helpful are the viewpoints organized and presented in the summary?'', specifically comparing our new form of claim-reason structured summary over the existing textual and KP summaries.
Note that to measure \textsc{Helpfulness}, we select summaries generated by the latest work from each form (e.g., \EQSUMM, ConvoSumm, and ConvPAKPA) to represent the 3 forms.

To perform this evaluation, we hire workers from Amazon Mechanical Turk (MTurk) to conduct pairwise comparisons of KPs from different systems.\footnote{We ensure inter-annotator consistency by selecting annotators with pairwise Cohen's Kappa $\geq 0.1$}.
Each comparison involved choosing the better one from two summaries, each taken from a different system.
Using the Bradley-Terry model~\citep{friedman-etal-2021-overview}, we calculated rankings from these comparisons among the models.

\paragraph{Results}
From Table~\ref{table:kp_quality_eval_results_T},
overall, \EQSUMM achieves consistent and up to 7.86 times improvements on all 7 dimensions, and are notably higher on \textsc{Coverage}, \textsc{Validity}, \textsc{Sentiment} and \textsc{Informativeness}.
In addition, ConvPAKPA achieves a slightly better score in \textsc{Sentiment} and \textsc{SingleAspect} than ConvoSumm thanks to its aspect-sentiment-based clustering approach to capture arguments.

Table~\ref{table:kp_quality_eval_results_T} further examines the \textsc{Helpfulness} of the claim-reason structured summary.
Overall, the claim-reason structured summary, produced by \EQSUMM (ours), GPT-4-ICL and QAF, yields up to 7 times more comprehesive and useful viewpoint presentation than other baselines.
In fact, while the traditional textual summary form (of ConvoSumm) already grounded generation with argument information for diversity, it failed to attach convincing evidence along viewpoints, nor presenting them in a logical manner.
Notably, the KP summary form, by producing a long and flat list of KPs, makes least sense of comprehension and usefulness as overlapping KPs (at different granularity) being scattered across the list inattentively.

\subsection{RQ2: How well the generated summaries represent arguments from the corpus?}
\paragraph{Settings}
This evaluation ignores argument structure and instead assesses the textual quality of arguments in generated summaries against the gold summaries. We aggregate all claims and reasons from each structured summary and reference summary as argument sets. Lexical similarity is computed using maximum $\mathtt{ROUGE}$ scores between generated and reference arguments. 
Then, following ~\citet{li-etal-2023-hear}, we calculate soft-Precision/Recall/F1~(denoted as sP, sR and sF1) to evaluate the \emph{semantic} similarity between individual generated argument and reference argument.
While $sP$ finds the reference argument with the highest similarity score for each generated argument, $sR$ is vice-versa, and ($sF1$) is the harmonic mean between $sP$ and $sR$.

\begin{equation}
  \small
  sP = \frac{1}{n} \times \sum_{ \alpha_i\in\mathcal{A}} \max_{\beta_j\in\mathcal{B}} f(\alpha_{i},  \beta_{j})
\end{equation}
\begin{equation}
  \small
  sR = \frac{1}{m} \times \sum_{ \beta_i\in\mathcal{B}} \max_{\alpha_j\in\mathcal{A}} f(\alpha_{i},  \beta_{j})
\end{equation}
where $f$ computes similarities between two individual key points, $\mathcal A$, $\mathcal{B}$ is the set of generated and reference KPs and $n=|\mathcal{A}|$ and $m=|\mathcal{B}|$, respectively.
We use state-of-the-art semantic similarity metrics $\mathtt{BLEURT}$~\citep{sellam-etal-2020-bleurt} and $\mathtt{BERTScore}$~\citep{Zhang*2020BERTScore}.
Additionally, we further measure the utility of argumentation theory in conversation summarization, by setting up a regular PAKPA that directly accepts user comments as input.

\paragraph{Results}
Table~\ref{table:automatic_evaluation} reports the lexical and semantic quality of generated argument in the summaries, 
with Reddit being the most challenging domain due to its informal language.
Overall, \EQSUMM consistently outperforms all baselines, achieving up to 3.71 times higher lexical similarity and a 0.22-point gain in semantic quality, thanks to disentangling the claim-reason clustering process in our argument-aware clustering.
This ultimately prevents reasons from being mixed within claim clusters, ensuring more specific and diverse justifications.
In contrast, QAF, which clusters propositions without role distinction, ranks second-lowest with weaker coherence (e.g., sF1 = 0.18 on Reddit). 
Nevertheless, our use of entailment scoring in \EQSUMM, rather than simple semantic similarity (e.g.,  QAF, also helps obtain a clearer distinction of subtle attitudes and stances, contributing to its superior sR (0.36 vs. 0.27 BLUERT sF1 on NYT).
Importantly, utilizing lightweight LLMs (e.g., \texttt{Mistral}) as backbone for \EQSUMM does not substantially degrades the textual quality due to our rigorous claim-reason clustering process.

Among the weaker baselines, KPA-based models (ConvPAKPA and ConvRKPA) perform poorly as they were not designed to process and summarize argument structure between comments'claims and reasons.
Nevertheless, ConvPAKPA outperforms QAF in Reddit and NYT, as its separate generation of claim and reason KPs, which help capture viewpoints more effectively, reflected in higher sR scores 
(0.34 vs. 0.27 BLUERT sF1 on NYT)

Finally, although GPT-4.1-ICL is a strong generative baseline, it underperforms \EQSUMM in both lexical quality (up to 19.3\%) and semantic quality (up to 22.9\%).
This is primarily due to hallucinations and the challenges LLMs face with long-context reasoning, 
where multi-step reasoning over complex propositions can make its performance less robust compared to \EQSUMM.

\subsection{RQ3: How well does the model match between the comments and the generated claims or reasons?}
\paragraph{Settings}
We evaluate how accurately each framework clusters original claims and reasons from input comments and matches them with those in the generated summary. 
We extend \citet{bar-haim-etal-2021-every} to measure both \emph{precision} (correctness of predicted matches) and \emph{recall} (coverage of ground-truth matches), 
by prompting \texttt{gpt-4-o-mini} to annotate pairwise \emph{match}/\emph{non-match} between generated and original claims or reasons.
Results are reported at three levels: claim, reason, and claim-reason.
The claim-reason level offers a stricter evaluation of the argument structure, requiring both a reason and its parent claim to be correctly matched to their respective summarized counterparts, thus evaluating structural consistency.
Empirical validation shows \texttt{gpt-4-o-mini} annotations highly correlated with MTurk workers' judgement (Pearson’s $r$ = 0.647).

\paragraph{Results}
Table~\ref{table:claim_reason_clustering_performance} reports how well different systems cluster claims and reasons in the final summary, evaluated at the claim, reason, and claim-reason levels.
\EQSUMM mostly achieves the highest precision, recall, and F1 across domains (e.g., 0.563 vs 0.142 Reason-Level F1 for Reddit)

Across all three domains, we also observe that \EQSUMM and \texttt{GPT-4.1}-ICL achieve highly stable, and sometimes higher, performance at the reason level compared to the claim level, largely due to their argument-aware clustering strategy. 
By first clustering claims and then aggregating their child reasons for reason-level clustering, these frameworks create more coherent reason groups. As a result, they achieve stronger structural consistency, which in turn boost the performance at the claim-reason level. (e.g., 0.728 F1 for \EQSUMM on Stack)
We also notice that limited computation of lightweight LLMs (e.g., \texttt{Mistral}) makes them score and cluster claims/reasons not as accurate as \texttt{GPT-4.1} in \EQSUMM, and even \texttt{GPT-4.1-ICL}.

In contrast, QAF, without disentangling the role of claims and reasons for clustering nor adopting hierarchical clustering,
shows much lower reason-level performance, especially in the NYT domain
(e.g., 0.168 F1 vs. 0.512 F1 of \EQSUMM).
In fact, with higher volume and complexity of arguments across NYT comments, reasons were more likely to be mixed in the claim cluster and therefore cannot contribute meaningfully to their truly expected reason clusters.

Similarly,
KPA-based approaches also underperform because they are not originally designed to model argument structure. 
However, ConvPAKPA still achieves better clustering performance than QAF thanks to separately processing claim and reason during clustering to avoid mixing.

\subsection{RQ4: How well and reasonable does the reason support its parent claim in an argument?}
\paragraph{Settings}
We further evaluate the convincingness of reasons supporting its parent claim along the structure in our generated summary leveraging claim verification task in fact checking.
Given a claim and every of its associated reason in the summary,
we prompt \texttt{gpt-4.1-mini} to annotate whether the reason \emph{supports} or \emph{refutes} the claim.
We then compute the precision of the \emph{support} label, which is the proportion of reasons judged as valid evidence for their associated claims.

\paragraph{Results}
Table~\ref{table:claim_verification_eval} shows that \EQSUMM consistently achieves the highest reason–claim support across all domains.
(e.g., 0.828 for \EQSUMM vs 0.738 for \texttt{GPT-4.1}-ICL on Stack).
This highlight the effectiveness of \EQSUMM’s entailment-based clustering and structured summary generation in maintaining logical coherence between claims and reasons.
GPT-4.1-ICL, while performing better than QAF, surpassed by \EQSUMM due to its one-step generation strategy, which can produce plausible but less rigorously justified reasons.
However, using light-weight LLMs for \EQSUMM cannot surpass GPT-4.1-ICL 
as these models cannot perform entailment-based scoring and clustering of claims/reasons as accurate as \texttt{GPT-4.1}.
QAF performs the weakest (0.568 on NYT), as its role-agnostic approach to clustering often leads to mismatched or loosely related reasons being associated with claims. 

\begin{table}[t]
    \centering
    \scalebox{0.6}{
    \begin{tabular}{lcccccccccc}
    \toprule
    {} & \multicolumn{3}{c}{ROUGE} & \multicolumn{3}{c}{BERTScore} & \multicolumn{3}{c}{BLEURT} \\
    \cmidrule(r){2-4} \cmidrule(r){5-7} \cmidrule(r){8-10}
     & R-1 & R-2 & R-L & sP & sR & sF1 & sP & sR & sF1 \\
    \midrule
    \rowcolor{lightgray}
    \multicolumn{10}{l}{$\mathbf{Reddit}$}\\
    \specialrule{0em}{1pt}{1pt}
    \EQSUMM (\texttt{GPT-4.1}) & \textbf{0.368} & \textbf{0.167} & \textbf{0.342} & 0.19 & \textbf{0.31} & \textbf{0.24} & 0.28 & \textbf{0.34} & 0.31 \\
    \EQSUMM (\texttt{Mistral}) & 0.357 & 0.151 & 0.335 & 0.19 & 0.28 & 0.23 & 0.32 & 0.33 & 0.33 \\
    \texttt{GPT-4.1}-ICL & 0.324 & 0.138 & 0.306 & 0.17 & 0.25 & 0.20 & 0.28 & 0.33 & 0.30 \\
    ConvoSumm & 0.337 & 0.128 & 0.312 & \textbf{0.26} & 0.22 & 0.24 & \textbf{0.36} & 0.31 & \textbf{0.33} \\
    QAF & 0.239 & 0.068 & 0.219 & 0.19 & 0.18 & 0.18 & 0.28 & 0.26 & 0.27 \\
    \hdashline
    ConvPAKPA & 0.282 & 0.070 & 0.257 & 0.23 & 0.27 & 0.24 & 0.30 & 0.26 & 0.28 \\
    ConvRKPA & 0.177 & 0.045 & 0.169 & 0.18 & 0.13 & 0.15 & 0.26 & 0.22 & 0.24 \\
    \rowcolor{lightgray}
    \multicolumn{10}{l}{$\mathbf{Stack}$}\\
    \specialrule{0em}{1pt}{1pt}
    \EQSUMM (\texttt{GPT-4.1}) & \textbf{0.480} & \textbf{0.275} & \textbf{0.447} & 0.24 & \textbf{0.35} & \textbf{0.28} & 0.38 & \textbf{0.42} & \textbf{0.40} \\
    \EQSUMM (\texttt{Mistral}) & 0.466 & 0.250 & 0.412 & 0.25 & 0.28 & 0.26 & 0.41 & 0.39 & 0.40 \\
    \texttt{GPT-4.1}-ICL & 0.439 & 0.222 & 0.404 & 0.21 & 0.27 & 0.24 & 0.35 & 0.40 & 0.38 \\
    ConvoSumm & 0.412 & 0.191 & 0.364 & \textbf{0.31} & 0.23 & 0.27 & \textbf{0.43} & 0.34 & 0.38 \\
    QAF & 0.402 & 0.173 & 0.366 & 0.26 & 0.22 & 0.24 & 0.41 & 0.36 & 0.38 \\
    \hdashline
    ConvPAKPA & 0.321 & 0.104 & 0.280 & 0.24 & 0.25 & 0.24 & 0.38 & 0.30 & 0.34 \\
    ConvRKPA & 0.256 & 0.084 & 0.242 & 0.23 & 0.13 & 0.17 & 0.34 & 0.26 & 0.29 \\
    \rowcolor{lightgray}
    \multicolumn{10}{l}{$\mathbf{NYT}$}\\
    \specialrule{0em}{1pt}{1pt}
    \EQSUMM (\texttt{GPT-4.1}) & \textbf{0.444} & \textbf{0.235} & \textbf{0.410} & 0.20 & \textbf{0.36} & \textbf{0.26} & 0.32 & \textbf{0.41} & \textbf{0.36} \\ 
    \EQSUMM (\texttt{Mistral}) & 0.428 & 0.213 & 0.391 & 0.21 & 0.32 & 0.25 & 0.33 & 0.38 & 0.35 \\
    \texttt{GPT-4.1}-ICL & 0.408 & 0.196 & 0.378 & 0.22 & 0.31 & 0.25 & 0.32 & 0.40 & 0.35 \\
    ConvoSumm & 0.371 & 0.165 & 0.342 & \textbf{0.30} & 0.23 & 0.26 & \textbf{0.39} & 0.30 & 0.34 \\
    QAF & 0.356 & 0.144 & 0.331 & 0.24 & 0.28 & 0.26 & 0.33 & 0.27 & 0.27 \\
    \hdashline
    ConvPAKPA & 0.365 & 0.137 & 0.334 & 0.25 & 0.30 & 0.27 & 0.36 & 0.32 & 0.34 \\
    ConvRKPA & 0.270 & 0.083 & 0.250 & 0.21 & 0.20 & 0.21 & 0.28 & 0.25 & 0.27 \\
    \bottomrule
    \end{tabular}
    }
    \caption{\label{table:automatic_evaluation}
    General textual quality of generated claim and reason KPs from the summary.
    }
\end{table}

\begin{table}[t]
    \centering
    \scalebox{0.55}{
    \begin{tabular}{lcccccccccc}
    \toprule
    {} & \multicolumn{3}{c}{\textbf{Claim Level}} & \multicolumn{3}{c}{\textbf{Reason Level}} & \multicolumn{3}{c}{\textbf{Claim-Reason Level}} \\
    \cmidrule(r){2-4} \cmidrule(r){5-7} \cmidrule(r){8-10}
     & P & R & F1 & P & R & F1 & P & R & F1 \\
    \midrule
    \rowcolor{lightgray}
    \multicolumn{10}{l}{$\mathbf{Reddit}$}\\
    \specialrule{0em}{1pt}{1pt}
    \EQSUMM (\texttt{GPT-4.1}) & \textbf{0.834} & \textbf{0.376} & \textbf{0.518} & \textbf{0.813} & 0.430 & \textbf{0.563} & \textbf{0.647} & \textbf{0.788} & \textbf{0.711} \\
    \texttt{GPT-4.1}-ICL & 0.771 & 0.251 & 0.379 & 0.626 & 0.375 & 0.469 & 0.513 & 0.701 & 0.592 \\
    \EQSUMM (\texttt{Mistral}) & 0.747 & 0.169 & 0.276 & 0.590 & 0.387 & 0.468 & 0.516 & 0.604 & 0.557 \\
    QAF & 0.683 & 0.157 & 0.255 & 0.581 & 0.265 & 0.364 & 0.470 & 0.581 & 0.519 \\
    \hdashline
    ConvPAKPA & 0.405 & 0.375 & 0.389 & 0.316 & \textbf{0.603} & 0.415 & -- & -- & -- \\
    ConvRKPA & 0.265 & 0.150 & 0.191 & 0.176 & 0.119 & 0.142 & -- & -- & -- \\
    \rowcolor{lightgray}
    \multicolumn{10}{l}{$\mathbf{Stack}$}\\
    \specialrule{0em}{1pt}{1pt}
    \EQSUMM (\texttt{GPT-4.1}) & \textbf{0.811} & \textbf{0.403} & \textbf{0.538} & \textbf{0.804} & 0.369 & \textbf{0.506} & \textbf{0.627} & \textbf{0.867} & \textbf{0.728} \\
    \texttt{GPT-4.1}-ICL & 0.730 & 0.147 & 0.245 & 0.610 & 0.319 & 0.419 & 0.572 & 0.626 & 0.598 \\
    \EQSUMM (\texttt{Mistral}) & 0.641 & 0.141 & 0.231 & 0.660 & 0.301 & 0.413 & 0.618 & 0.528 & 0.569 \\
    QAF & 0.618 & 0.130 & 0.214 & 0.750 & 0.241 & 0.364 & 0.417 & 0.562 & 0.479 \\
    \hdashline
    ConvPAKPA & 0.680 & 0.233 & 0.347 & 0.357 & \textbf{0.565} & 0.438 & -- & -- & -- \\
    ConvRKPA & 0.500 & 0.125 & 0.200 & 0.533 & 0.072 & 0.127 & -- & -- & -- \\
    \rowcolor{lightgray}
    \multicolumn{10}{l}{$\mathbf{NYT}$}\\
    \EQSUMM (\texttt{GPT-4.1}) & 0.573 & \textbf{0.558} & \textbf{0.566} & \textbf{0.756} & 0.387 & \textbf{0.512} & 0.485 & \textbf{0.837} & \textbf{0.614} \\
    \texttt{GPT-4.1}-ICL & \textbf{0.827} & 0.188 & 0.307 & 0.691 & 0.298 & 0.416 & \textbf{0.592} & 0.604 & 0.598 \\
    \EQSUMM (\texttt{Mistral}) & 0.523 & 0.212 & 0.302 & 0.656 & 0.282 & 0.394 & 0.365 & 0.627 & 0.461 \\
    QAF & 0.517 & 0.144 & 0.225 & 0.367 & 0.109 & 0.168 & 0.250 & 0.050 & 0.080 \\
    \hdashline
    ConvPAKPA & 0.568 & 0.326 & 0.414 & 0.427 & 0.585 & 0.494 & -- & -- & -- \\
    ConvRKPA & 0.431 & 0.264 & 0.327 & 0.257 & 0.111 & 0.155 & -- & -- & -- \\
    \specialrule{0em}{1pt}{1pt}
    \bottomrule
    \end{tabular}
    }
    \caption{\label{table:claim_reason_clustering_performance}
    Comment matching correctness of claim and reason KPs in the generated summary, measured at different level. 
    Claim-Reason Level not applicable for ConvPAKPA and ConvRKPA 
    as they were not designed to process and summarize argument structure between claims and reasons.
    }
\end{table}

\begin{table}[!ht]
    \centering
    \scalebox{0.65}{
    \begin{tabularx}{0.48\textwidth}{|l|X|X|X|}
    \hline
        & $\mathbf{Reddit}$ & $\mathbf{Stack}$ & \textbf{NYT} \\ \toprule
        \EQSUMM (\texttt{GPT-4.1}) & \textbf{0.804} & \textbf{0.828} & \textbf{0.764} \\
        \texttt{GPT-4.1}-ICL & 0.687 & 0.738 & 0.612 \\
        \EQSUMM (\texttt{Mistral}) & 0.669 & 0.710 & 0.594 \\
        QAF & 0.652 & 0.675 & 0.568 \\ \bottomrule
    \end{tabularx}
    }
    \caption{\emph{Precision} of reason-claim support (convincingness) from the generated summary. Applicable only to baselines outputting claim-reason structured summary.}
    \label{table:claim_verification_eval}
\end{table}

\section{Conclusion}
In this paper, we introduced \EQSUMM, a novel task and framework of argument-aware quantitative summarization for online discussions 
which generate structured summaries composed of claims and their supporting reasons. 
Different from previous works, our approach leverages entailment-based clustering to disentangle claims and reasons, followed by hierarchical aggregation to preserve both specificity and diversity of viewpoints. 
We further designed a structured generation strategy to ensure logical alignment between claims and reasons, addressing the limitations of existing textual summarization methods.
Experiments on multiple forms of conversation social comments demonstrate that \EQSUMM consistently delivers higher lexical and semantic quality, better clustering performance, and more convincing reasoning compared to baselines.

\section*{Acknowledgement}
This research is supported in part by the Australian Research Council Discovery Project \textbf{DP200101441}.

\bibliography{aaai2026,anthology_0,anthology_1,custom}

\end{document}


\maketitle

\appendix

\section{Human Validation of the LLM-based Entailment Scoring Method in \EQSUMM}
\subsection{Settings}
To validate the effectiveness of our LLM-based entailment scoring method applied during argument-aware clustering in \EQSUMM, we measure how well method's results align with human judgement, compared with the alignment made by state-of-art-model \texttt{RoBERTa-large} model~\footnote{\url{https://huggingface.co/roberta-large-mnli}} on our task.
We specifically conducted this experiment by sampling and transforming our LLM-based's entailment score (from 30 pairs per domain) into 3-way MNLI labels (entailment (1-2.5), neutral (2.5-3.5), and contradict (3.5-5)).
For \texttt{RoBERTa-large}, we select the class 
with highest probability as the final label for each pair.
We then employed 4 MTurk crowd workers on every proposition pair for labelling 
entailment correctness,
in which at least 60\% of the annotators had to agree that the match is correct, otherwise, it is incorrect.
For quality control, we select only annotators with an 80\% or higher approval rate and at least 10 approved tasks. 
Following \citet{bar-haim-etal-2021-every}, we exclude annotators with Annotator-$\kappa <0.1$ for quality control.
This score averages all pairwise Cohen's Kappa~\citep{landis1977measurement} for a given annotator, for any annotator sharing at least $10$ judgments with at least $3$ other annotators.



\begin{table}[htbp]
  \centering
  \small
  \begin{tabular}{@{}p{2.5cm}cc@{}}
    \toprule
    Model & Acc. & $r$ \\
    \midrule
    \texttt{GPT-4.1}       & 0.715 & 0.556 \\
    \texttt{RoBERTa-large} & 0.560 & 0.175 \\
    \bottomrule
  \end{tabular}
  \caption{Human-validated entailment scoring. Accuracy (Acc.) and Pearson’s $r$.}
  \label{table:llm_entailment_annotation_validate}
\end{table}

\subsection{MTurk Annotation Guideline}
Below are the annotation guidelines for MTurk workers to annotate a given proposition (claim or reason) pair:\\

In this task you are presented an a topic from online discussion thread (e.g., a community question, topic), claim A and claim B extracted from the social comments answering the topic.
You will be asked to answer the following question:
"Does claim A support claim B?"\\

\textbf{Question:} one more 980ti or wait for new nvidia cards? so sli 980ti or wait for new nvidia gpus?

\textbf{Claim A:} Used ones will be a lot cheaper as early adopters dump theirs

\textbf{Claim B:} You should buy the best one when it comes out\\

The options are:
\begin{itemize}
\item Contradiction
\item Neutral
\item Entailment

\end{itemize}

\subsection{Results}
From Table~\ref{table:llm_entailment_annotation_validate}, we observed that prompting \texttt{GPT-4.1} for measuring entailment significantly outperforms the \texttt{RoBERTa-large} baseline in both accuracy and correlation with human annotations. Specifically, the Pearson’s $r$ of 0.556 suggests that 
our LLM-based approach
produces scores that are
correlates relatively strong with
human judgment for entailment, 
compared to only 0.175 from \texttt{RoBERTa-large}. This finding substantiates our use of \texttt{GPT-4.1} as the backbone for entailment-based claim-reason clustering in \EQSUMM.

\section{Details of Summary Information Quality and Utility Evaluation}
\label{sec:kp_quality_dimensions}
This section provides detailed descriptions of the tasks and 8 criteria involved in our manual evaluation of \emph{the utility of the argument structure} in the summaries, and also the summary's \emph{information quality}.
While information quality is adopted from the 7 different criteria, 
(e.g., \textsc{Faithfulness}, \textsc{Coverage})
defined by previous KPA studies~\citep{kapadnis-etal-2021-team}, 
we additionally define \textsc{Helpfulness} to evaluate `` how helpful are the viewpoints organized and presented in the summary?'', specifically comparing our new form of claim-reason structured summary over the existing textual and KP summaries.
Note that to measure \textsc{Helpfulness}, we select summaries generated by the latest work from each form (e.g., \EQSUMM, ConvoSumm, and ConvPAKPA) to represent the 3 forms.

Annotators were asked to perform a pairwise comparison between two sets of summaries, each taken from a different model, generated for a specific 
discussion thread.
The annotators must answer a comparative question with respect to the evaluating criteria. (e.g., \emph{Which of the two summaries captures better [Criterion]?}).
For each criterion, following~\citet{friedman-etal-2021-overview}, we calculate the ranking using the Bradley-Terry model~\citep{bradley_terry}, which predicts the probability of a given participant winning a paired comparison, based on previous paired comparison results of multiple participants, and thus allows ranking them.

For annotation setting, we employed 4 MTurk crowd workers on every comparison sample for labelling the better summary,
in which at least 60\% of the annotators had to agree that the match is correct, otherwise, it is incorrect.
For quality control, we select only annotators with an 80\% or higher approval rate and at least 10 approved tasks. 
Following \citet{bar-haim-etal-2021-every}, we exclude annotators with Annotator-$\kappa <0.1$ for quality control.
This score averages all pairwise Cohen's Kappa~\citep{landis1977measurement} for a given annotator, for any annotator sharing at least $20$ judgments with at least $3$ other annotators.

\subsection{Evaluation Criteria Description}
The description of 8 evaluation criteria is as follows:



\begin{itemize}
    \item \textsc{Validity}: The arguments in the summary should be presented as understandable, well-written sentences or phrases representing an opinion of the users towards the topic. This would filter out sentences such as \emph{``It's rare these days to find that!''}
    \item \textsc{Sentiment}: The arguments in the summary should have a clear sentiment towards the product being questioned (either positive or negative). This would exclude sentences like 'I was introduced this product by a friend'.
    \item \textsc{Informativeness}: 
    The arguments in the summary should discuss aspects relating to the discussing topic and contain useful information. 
    Any argument that is too general or only expresses sentiment cannot be considered a good candidate. Statements such as 'Love this idea' or 'We were very disappointed', which merely express an overall sentiment, should be discarded.    
    \item \textsc{SingleAspect}: The arguments in the summary should not discuss multiple aspects (e.g., 'The bike is very sturdy and the price is reasonable.' is not a good viewpoint).")    
\end{itemize}

\begin{itemize}
    \item \textsc{Redundant}: 
    Each argument in the summary should express a distinct aspect. In other words, there should be no overlap between the arguments.
    \item \textsc{Diversity}: 
    The arguments in the summary should cover a wide diversity of opinions relevant and representative to the topic.
    \item \textsc{Faithfulness}: 
    The arguments in the summary should express reasonable and meaningful opinions relevant to the topic without hallucination. No conjecture or unfounded claims should arise.
    \item \textsc{Helpfulness}:
    The delivery and presentation of viewpoints in the summary should be practically useful and comprehensible for real users to understand the discussion content.    
\end{itemize}

\subsection{Annotation Guideline}
Below are the annotation guidelines for given summary pairs representing two baselines from our experiment:\\

Below are the two summaries for a topic (e.g., online question, news article) from an online discussion thread, generated by two different summarization frameworks. Each summary, possibly generated in different forms, contains several key points (i.e., salient points) generated by summarizing user opinions from user's social comments on different aspects in answering the topic. You are tasked to select which summary you think is better according to the below criteria.\\

\textbf{Domain:} News Comments (New York Times).

\textbf{Topic:} Wednesday: Keeping up with the cold, a verdict in the Etan Patz case, and the 5-cent plastic bag fee..

\textbf{Criteria:} \textsc{Helpfulness}. The delivery and presentation of viewpoints in the summary should be practically useful and comprehensible for real users to understand the discussion content.\\

\textbf{Summary A:}\\
\underline{Claim: Vaccination protects you and others}\\
\hspace*{0.1em} $\rightarrow$ \textbf{Reason:} \textit{Vaccination is important for everyone, especially to protect vulnerable groups like the elderly, young children, and those with chronic conditions, even though effectiveness varies and not getting sick at all is ideal.}\\
\hspace*{0.1em} $\rightarrow$ \textbf{Reason:} \textit{Vaccination is especially important for those in or around high-risk groups, such as the elderly or immunocompromised, to prevent illness and protect the community.}\\
\underline{Claim: Governor's concern is misplaced}\\
\hspace*{0.1em} $\rightarrow$ \textbf{Reason:} \textit{Governor should focus on bigger issues like housing and childcare.}\\
\hspace*{0.1em} $\rightarrow$ \textbf{Reason:} \textit{Plastic bag fee is affordable for most people.}\\
\underline{Claim: Reuse plastic bags as garbage bags}\\
\hspace*{0.1em} $\rightarrow$ \textbf{Reason:} \textit{The 5¢ plastic bag fee encourages people to reuse bags.}\\
\hspace*{0.1em} $\rightarrow$ \textbf{Reason:} \textit{Reusing plastic bags saves money on buying garbage bags.}\\

\textbf{Summary B:} Several commenters discuss the importance of getting vaccinated against the flu. One commenter says that it is important to have the right vaccine, but another says that this is not always the case. A couple of commenters talk about the need for plastic bags to be recyclable or biodegradable in order to protect the environment.\\

The options are:
\begin{itemize}
\item Summary A
\item Summary B
\end{itemize}

\section{Prompts in \EQSUMM}

\subsection{Prompts for argument proposition extraction}
We present the zero-shot prompt for argument proposition extraction from comments according to the Toulmin Theory in Listing~\ref{lst:gpt_arg_ext_0_shot}.

\begin{figure*}
  \centering
\begin{lstlisting}[caption={Zero-shot prompt for Argument Proposition Extraction using \texttt{gpt-4.1}.}, basicstyle=\small, label=lst:gpt_arg_ext_0_shot]
Help me extract the proposition of the given comment according to Toulmin model. Extracted proposition should be original sentences/phrases from the comment. You can ignore non argumentative unit/senteces. Every proposition must refer to a specific object/subject the subject and cannot be pronoun (e.g., it, this, that):

Extract all possible claims and their propositions as a list of JSONs:
[{'claim': <claim>, 'ground': [<list of grounds>], ...}, ...]

Comment: "While I admire your steadfast courage in how awesome Girafig is, I may wait on that one. :P BUT. Maybe. :)"
According to Toumin theory,
\end{lstlisting}
\end{figure*}

\subsection{Prompts for entailment scoring}
We present the zero-shot prompt for scoring the entailment between argument proposition during argument-aware clustering in Listing~\ref{lst:gpt_nli_score_0_shot}.

\begin{figure*}
    \centering
\begin{lstlisting}[caption={Zero-shot prompt for Entailment Scoring using \texttt{gpt-4.1}.}, basicstyle=\small, label=lst:gpt_nli_score_0_shot]
You will be given a community question, a claim **A** and a list of other claims **B** extracted from the social comments answering the question.

Your task is to assess the **degree to which claim **A** supports each claim from **B** list, using a scale from 1 to 5.

Please make sure you read and understand these instructions carefully. Please keep this document open while reviewing, and refer to it as needed.

Evaluation Criteria:
Entailment Strength (1 to 5): how strongly the **claim A** logically supports the other claim.

Evaluation Steps:

1. Read both the claim **A** and each claim from list **B** carefully.
2. Determine if claim **A** logically follows each claim from list **B**.
3. For each claim from list **B**, assign an **entailment strength score from 1 to 5** by claim **A** according to the guideline above.
4. Return all scores as a Python list
\end{lstlisting}
\end{figure*}

\subsection{Prompts for claim-reason structured summary generation}
We present the zero-shot prompt for generating the final claim-reason structured summary, given claim clusters and their supporting reason clusters, in Listing~\ref{lst:gpt_structured_summ_gen_0_shot}.

\begin{figure*}
  \centering
\begin{lstlisting}[caption={Zero-shot prompt for prompting \texttt{gpt-4.1} on claim-reason structured summary generation.}, basicstyle=\small, label=lst:gpt_structured_summ_gen_0_shot]
In this task you are presented with a community question, a JSON object containing argument propositions (list of claims and their grounds) extracted from social comments answering the question.
The question asks the opinions of users, and can be generally answered by the list of propositions under 'claim'. The 'claim' can then be further justified by lists (clusters) of propositions under 'ground'

You were tasked to generate the general opinion implied by propositions under 'claim' to answer the question, and also generate every ground supported by each list of propositions under 'ground'

Perform the following actions to solve this task:
+ Generate a concise key point (KP) that captures salient opinions from the list of propositions under 'claim' --> 'Claim KP'
+ Then, for every list of propositions under 'ground', generate a concise key point that captures salient opinions across the propositions that support the 'Claim KP' --> 'Ground KP'

Note that the claim and ground must have less than 10 tokens. The claim should be general and the ground should be specific and cover details as much as possible, and the ground must strongly support the claim.
\end{lstlisting}
\end{figure*}

\subsection{Prompts for annotating the matching correctness between the
comments and the generated claims or reasons (RQ3)}
We present the our prompt for instructing \texttt{gpt-4-o-mini} to annotate pairwise \emph{match}/\emph{non-match} between generated and original claims or reasons in Listing~\ref{lst:gpt_claim_reason_comm_match_eval_0_shot}.

\begin{figure*}
  \centering
\begin{lstlisting}[caption={Zero-shot prompt for \texttt{gpt-4.1-mini} to annotate pairwise \emph{match}/\emph{non-match} between generated and original claims or reasons in RQ3}, basicstyle=\small, label=lst:gpt_claim_reason_comm_match_eval_0_shot]
In this task you are presented with a community question, a claim or reason taken from the summary answering the question, and a social comment answering the question.
You will be asked to answer the following question: "Does the claim or reason match, i.e, represent an opinion in the comment?"
A comment might express opinions on multiple aspects. A claim or reason only matches a comment if they explicitly discusses issues relating to each other.

The options are:
- Not At All
- Somewhat Not Well
- Somewhat Well
- Very Well

Remember to explain your answer in the output.

Community Question: ...
Claim or Reason: ...
Comment: ...
\end{lstlisting}
\end{figure*}

\subsection{Prompts for annotating the reasonability of a reason in supporting its parent claim (RQ4)}
We present the our prompt for instructing \texttt{gpt-4-o-mini} to annotate 
whether the reason \emph{supports} or \emph{refutes} the claim in Listing~\ref{lst:gpt_claim_reason_reasonability_eval_0_shot}.

\begin{figure*}
  \centering
\begin{lstlisting}[caption={Zero-shot prompt for \texttt{gpt-4.1-mini} to annotate  the reasonability of a reason in supporting its parent claim in RQ4}, basicstyle=\small, label=lst:gpt_claim_reason_reasonability_eval_0_shot]
base_prompt = """In this task you are presented with a community question, a claim and evidence extracted from the summary answering the question.
The question asks the opinions of users, and can be generally answered by the list of comments under 'claim'. The 'claim' can then be further justified by lists (clusters) of comments under 'ground'

You were tasked to verify if the evidence support the claim in answering the query.
You will first explain your reasoning, and then make the final judgment by writing LABEL: followed by a single word from the possible labels ['SUPPORTS', 'REFUTES'].

Question: ...
Claim: ...
Evidence: ...
\end{lstlisting}
\end{figure*}

\bibliography{aaai2026,anthology_0,anthology_1,custom}